\documentclass[letterpaper]{article}
\usepackage{flairs}
\usepackage{times}
\usepackage{helvet}
\usepackage{courier}
\usepackage{graphicx}
\graphicspath{ {figures/} }
\usepackage{caption}
\usepackage{subcaption}
\usepackage{wrapfig}
\usepackage{multirow}
\usepackage{tabularx}
\usepackage{paralist}
\usepackage{colortbl}
\definecolor{Gray}{gray}{0.9}

\usepackage{array} 

\begin{document}
%
\title{Malicious or Benign?\\ Towards Effective Content Moderation for Children's Videos}
\author{
\begin{tabular}{llll}
Syed Hammad Ahmed$^1$ & Muhammad Junaid Khan$^1$ & H. M. Umer Qaisar$^2$ & Gita Sukthankar$^1$\\
\end{tabular}
\\
$^1$Department of Computer Science, University of Central Florida, Orlando, FL USA\\
$^2$Royal Cyber,  Karachi, Pakistan\\
\{hammad.ahmed,junaid\_k\}@knights.ucf.edu \quad hafiz.qaisar@royalcyber.com \quad gitars@eecs.ucf.edu\\ 
}

\maketitle

\begin{abstract}
Online video platforms receive hundreds of hours of uploads every minute, making manual content moderation impossible.   
Unfortunately, the most vulnerable consumers of malicious video content are children from ages 1-5 whose attention is easily captured by bursts of color and sound.   Scammers attempting to monetize their content may craft malicious children's videos that are superficially similar to educational videos, but include scary and disgusting characters, violent motions, loud music, and disturbing noises.   Prominent video hosting platforms like YouTube have taken measures to mitigate malicious content on their platform, but these videos often go undetected by current content moderation tools that are focused on removing pornographic or copyrighted content.  This paper introduces our toolkit (\textbf{M}alicious \textbf{o}r \textbf{B}enign) for promoting research on automated content moderation of children's videos.
We present 1) a customizable annotation tool for videos, 2) a new dataset with difficult to detect test cases of malicious content and 3) a benchmark suite of state-of-the-art video classification models. 
\end{abstract}

\section{Introduction}
Online Video Platforms (OVP) facilitate the sharing of videos from a wide variety of genres; YouTube and TikTok are currently the top video hosting platforms with over 2.2 and 1 billion active users, respectively \cite{doyle_2022_tiktok}. In addition to these dedicated OVPs, social networks like Facebook, Twitter, and Instagram also allow users to post videos. According to \citeauthor{ruby_2022_youtube} (\citeyear{ruby_2022_youtube}), 720,000 hours of video are uploaded to YouTube daily to be consumed by users from a wide range of demographics including children under 12.
Children's videos can be prime targets for content creators seeking to monetize their work; out
 of the ten most watched videos on YouTube, five are children's cartoon videos \cite{ceci_2022_youtube}. One of these videos was disliked by many parents who considered it to be ``hypnotizing'' and ``creepy'' \cite{sinelschikova_2020_this}. 
 
There are many YouTube channels, such as Number Blocks, that use cartoon characters to deliver tailored educational content in order to improve the communication and reasoning skills of children ages 1-5.  By designing videos that are superficially similar in style, scammers can get their videos recommended based based on similarity of titles, hashtags, thumbnails, or meta-data.  \textit{Malicious} videos intentionally include audiovisual features which captivate the viewer's attention such as rapid repetitive motion, disgusting and scary characters, disturbing noise, and loud sounds. \citeauthor{10} (\citeyear{10}) note that a child who selects a recommended malicious video is more prone to continue watching a series of similar videos based on the video recommender system. This problem is of grave concern since 1) these children are unable to distinguish the appropriateness of the video content by themselves and 2) their parents are unable to constantly supervise their viewing \cite{screen_time}.

Overexposure to malicious videos is a significant risk factor for the cognitive development of preschool children.  Psychological studies conclude that overexposure to cartoons with fast motions may lead to a deterioration in the performance of commonplace tasks \cite{Lillard_Peterson_2011}. Similarly, exposure to loud noises inhibits a child's brain development especially reading, writing and comprehension skills \cite{Klatte_Bergström_Lachmann_2013}. Violent cartoons can directly affect the behavior of pre-school children, causing aggression and anxiety \cite{violent_cartoons}. 
An emerging genre of video content actively posted on OVPs include video game walkthroughs and gameplays. Although these are not explicitly malicious, gameplay videos often contain similar fast-motion and violent activities which may jeopardize the cognitive development of preschoolers.

\begin{figure}[htp]
    \centering
    \includegraphics[width=0.8\columnwidth]{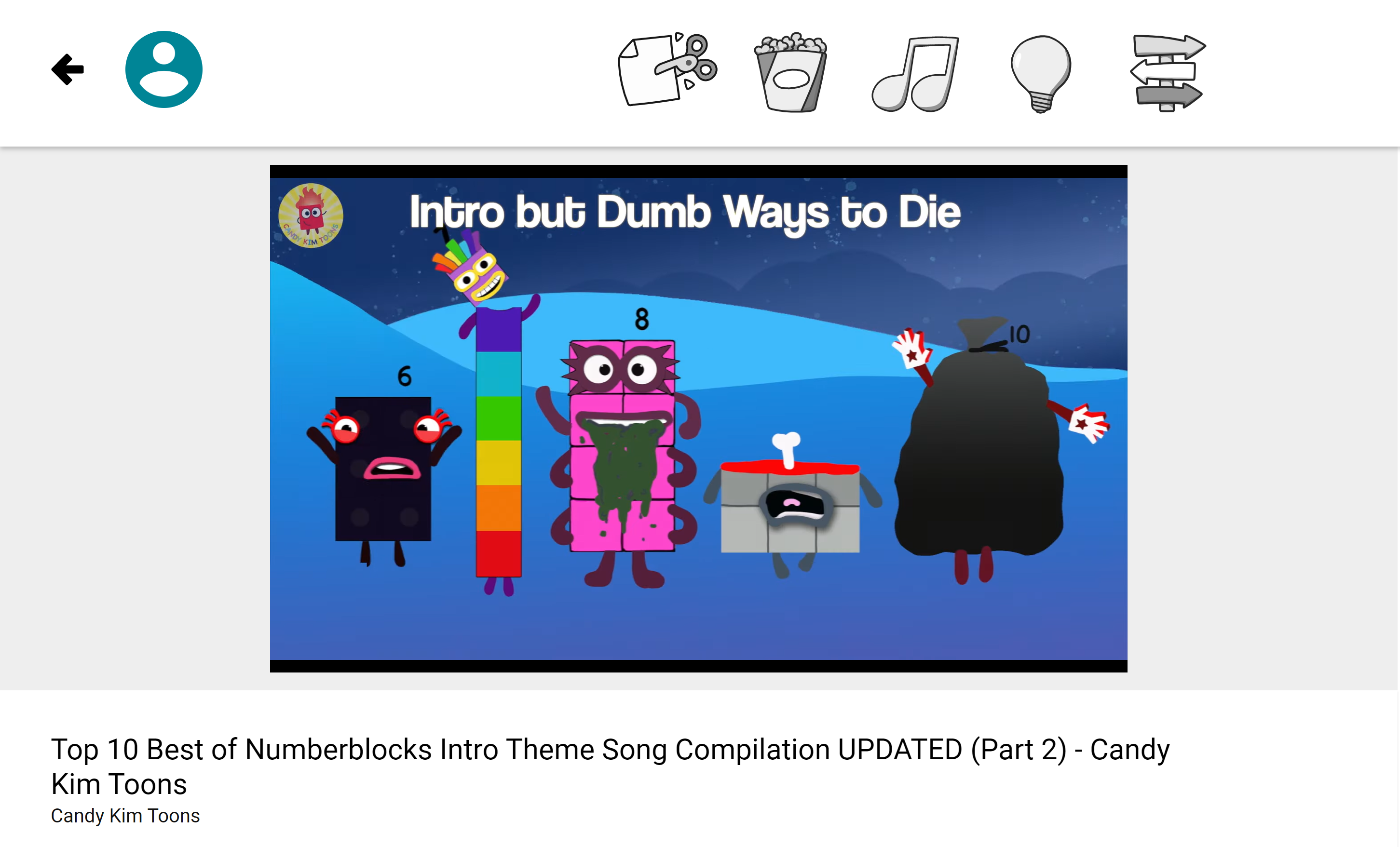}
    \caption{An example of a malicious video still viewable on YouTube Kids that is superficially similar to Number Blocks.}
    \label{fig:example}
\end{figure}

\nocite{federaltradecommission_2020_complying}
YouTube is trying its best to mitigate the availability of these videos to kids.  They introduced the ``made for kids'' flag which must be set by the video uploader in compliance with the Children’s Online Privacy Protection Act (COPPA).  YouTube also checks uploaded video content using AI techniques to moderate the videos.  Although many malicious videos are automatically removed by the YouTube automation, there have been instances where the platform overruled the creator's flag and mistakenly set videos as having been made for kids even when marked as inappropriate by the creator \cite{amadeo_2022_youtube}. Only videos marked as ``made for kids'' can be viewed on the kid-friendly application launched by YouTube called YouTube Kids (YTK); although this is very useful, malicious content remains viewable for long periods of time on YTK. 
 Figure~\ref{fig:example}  shows an example of a video still available on YTK that includes violent behaviors executed by the Number Blocks characters.

A variety of approaches have been used to tackle this problem including leveraging user comments \cite{7}, metadata \cite{1,6} and video features \cite{3,9,11}; others have employed multimodal techniques rather than relying on a single data source \cite{2,6,9,10}.  We focus on analyzing and annotating subtle audiovisual features present in cartoon videos which are not conducive for the overall mental and physical development of toddlers and pre-school kids. 
This paper makes the following contributions:
\begin{compactenum}
    \item Identifying a set of malicious features that are harmful to preschool children but remain unaddressed.
    \item Publishing a labeled dataset (\textbf{M}alicious \textbf{o}r \textbf{B}enign) of cartoon video clips. For each video clip, the presence or absence of each feature was marked by the annotators. We will make our MOB dataset publicly available to stimulate further research in this area.
    \item Providing a user-friendly web-based video annotation tool which can easily be customized and used for video classification tasks with any number of ground truth classes. 
    \item A thorough analysis of baseline results for a  few SOTA video classification benchmarks.
\end{compactenum}

\section{Background}\label{BG} 
\subsection{Age Groups of Children}
We assume that children have an age range from newborn to 17 years, after which one is considered as an adult. Children's age groups should be clearly defined when considering appropriate or inappropriate content. In terms of the content available on OVPs, a video considered appropriate for a specific age group may not be suitable for another, and vice versa.  Age groups are defined as follows: infants (newborn-1 year old), toddlers (1-3 years), pre-schoolers (4-5 years), middle childhood (6-11 years), and teenagers (12-17 years old) \cite{cdc_2019_positive}. The age groups most vulnerable to malicious video content are toddlers and preschoolers, whom we focus on in this paper.

\subsection{Malicious or Benign Classification}
\subsubsection{Malicious}
This paper assumes that video content which may not be suitable for viewing by toddlers and pre-schoolers includes a set of clearly defined, trivial and intuitive features as well as some complex and subtle audio and video features. The former includes forms of obscenity and violence on a higher level e.g. nudity, gore, etc. while the latter includes elements such as fast repetitive motion, loud music, disgusting and scary characters, smashing people or things, forms of aggression, loud music, screaming or shouting, gunshots and explosions. Furthermore, most anime are made for kids 10 and above, but a few anime series which are widely viewed by kids may contain ``fantasy violence''. 
 Similarly, certain video game gameplays are not appropriate as they may contain physical violence, strong language, drug and substance use, etc.  Videos which depict famous cartoon characters dressed inappropriately and performing strange and obscene activities, as were propagated during Elsagate, are also considered unsuitable for toddlers and pre-schoolers, and possibly for older kids as well.  
\subsubsection{Benign} 
Educational videos and videos of nursery rhymes are usually considered to be appropriate for the toddlers and pre-schoolers; in fact some experts recommend letting kids watch them for a limited number of hours \cite{cdc_2019_positive}.  Benign videos are characterized by a slower tempo, softer music or sound effects, moderate-paced motions, and soft-toned conversations.  Most importantly, benign video content should not contain any indicators of malicious content as discussed earlier in this section. The absence of all malicious features in a video qualifies it to be marked as benign by our annotators.

\subsection{Privacy and Rating Standards}
\subsubsection{Children's Online Privacy Protection Act (COPPA)} US Federal Trade Commission's (FTC) COPPA clearly defines strict privacy and content type guidelines for online content publishers specifically for 13-year-old kids or younger. In 2019, COPPA added more clauses specifically for YouTube to ensure kids watch safe content. 
\subsubsection{Video Content Rating Standards} 
\begin{figure*}[htp]
    \centering
    \includegraphics[width=0.8\columnwidth]{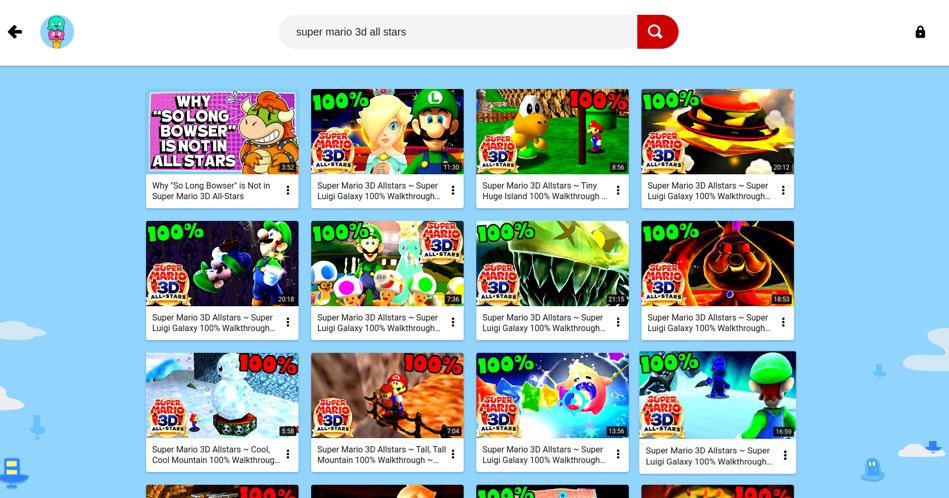}
    \caption{Gameplay videos for Super Mario 3D All-Stars which is rated as inappropriate for kids below 7 years of age.}
    \label{fig:ytkpegi}
\end{figure*}

The Motion Picture Association (MPA) regulates movie content ratings. It provides a comprehensive rating system where the ``G'' (general audience) rating category is considered suitable for kids. Similarly, for video games, the ESRB (Entertainment Software Rating Board) provides well-defined rating categories and content features. According to ESRB, the appropriate video game content for kids younger than 5 years is the ``E'' (everyone) category. Pan-European Game Information (PEGI) is another such regulatory body devoted to video game ratings. As an important genre of videos on OVPs is video game gameplays and walkthroughs, we have included their rating guidelines to help us identify malicious videos. The YouTube Kids platform includes video game gameplay videos for games which are rated as not appropriate for kids according to these rating standards. Figure \ref{fig:ytkpegi} (Appendix) shows the snapshot of search results of gameplay videos for Super Mario 3D All-Stars on YouTube Kids. Although it is rated PEGI-7 by PEGI for including frightening or violent content, it is still viewable on YouTube Kids platform for kids of all age groups.

\section{Related Work}\label{RW}
Many solutions have been proposed to the problem of detecting video content that is inappropriate for children. A general dichotomy between these techniques is whether they rely on language classification models and leverage video meta-data, user comments, and sub-titles as inputs \cite{1,6,7},  use the video stream alone \cite{3,9,11}, or use a multimodal approach that  combines both vision and language cues \cite{2,5,8,10}.

Only a small set of prior work has focused on cartoon videos \cite{2,10,11}, whereas most other authors have  studied videos with human characters, movie clips, news, and sports.  Elsagate \cite{elsa}, the misuse of popular cartoon characters to create clickbait content, triggered a surge of interest in content moderation for cartoons.
\citeauthor{11} (\citeyear{11}) proposed a fusion model where they extract features from frames and the related motion vectors. They focus specifically on Elsagate videos and have published a related dataset.

Recurrent neural networks are a popular approach for identifying inappropriate videos, and we utilize an LSTM as a benchmark. \citeauthor{3} (\citeyear{3}) classify cartoon videos as safe or unsafe using an LSTM based model with around 96\% accuracy.  KidsGUARD \cite{9} used an LSTM autoencoder to classify four classes of video:  violent, sexual, both, and safe.  

There is no single labeling strategy that is uniformly agreed upon for classifying inappropriate content.  
\citeauthor{6} (\citeyear{6}) proposed a meta-data based solution for identifying inappropriate videos. They presented results for both binary and multi-class scenarios in which the latter consists of four classes: suitable, disturbing, restricted, and irrelevant. In their work, gaming videos were marked as irrelevant whereas it is an important genre of kids videos which may contain inappropriate content.  Our dataset also includes these gameplay videos.

\citeauthor{1} (\citeyear{1}) focus on a slightly different problem: instead of identifying malicious videos they focus on detecting channels which post malicious videos.  An inherent assumption in their work is that if even one disturbing video is detected, all videos posted by that channel are flagged as disturbing, making it vulnerable to false positives. 

\citeauthor{5} (\citeyear{5}) introduced the Samba framework which is a multimodal solution that uses subtitles as input in addition to video metadata. Since neither music nor sound effects are depicted in the subtitles, a video which has malicious audio content  will still be classified as safe by Samba. We manually checked a few videos from the Samba dataset and found that most of the inappropriate content was not cartoons and included news, Hollywood movies, and TV shows.

\citeauthor{cartoonsurvey} (\citeyear{cartoonsurvey}) performed a comprehensive survey on the impact of cartoons on kids; they found that children are more enticed by cartoon characters vs.\ non-cartoon characters. Since less attention has been devoted to cartoons, we focus on cartoon videos only and have collected a  dataset of malicious cartoon videos that are difficult to detect.
\citeauthor{2} (\citeyear{2}) proposed a multimodal solution using text (user comments and captions), along with image data.  As images do not capture the motion related features it is likely that malicious spatio-temporal patterns will be ignored by their method. 

\citeauthor{4} (\citeyear{4}) try to address the problem of finding inappropriate advertisements shown on appropriate videos. They only present a statistical analysis of the data gathered and did not implement a machine learning classifier.
The analysis performed by \citeauthor{10} (\citeyear{10}) reveals that malicious content creators have strong connections among each other and they promote each other's contents by commenting, liking, and sharing playlists.
\citeauthor{8} (\citeyear{8}) propose a solution which distinguishes between original and fake cartoon videos.  They also further identify the malicious fake cartoons as being violent or explicit. 

To summarize, this paper tries to address the following general limitations in previous work:
 1) no psychological studies were used in prior work to support why the content was unsuitable for kids of ages 1-5; 2) annotators were directly asked about their opinion whether the videos are appropriate or inappropriate for young children which makes the dataset more subjective; 3) spatio-temporal audiovisual features that are not conducive to children's mental and physical health are omitted from the malicious examples in other datasets.  We hope that our annotation tool, dataset, and benchmarks will be a helpful resource for authors seeking to benchmark their video content moderation approaches.

\section{Methodology}\label{M}
\subsection{The MOB Dataset}\label{DS}
Our dataset\footnote{https://github.com/syedhammadahmed/mob} focuses on cartoon videos which contain activities or events not suitable for viewing by children of ages 1-5 (toddlers and pre-schoolers); examples include rapid repetitive motions, scary or disgusting characters, violent actions (e.g. hitting, smashing, biting, kicking),  obscene appearances, loud music or noise, screaming, shouting, explosion and gunshot sounds, and use of any offensive language.  Our dataset contains a more diverse set of cartoon videos and video game gameplays than has been previously publicly released.
\subsubsection{Relevant Datasets}
There are only a few existing datasets suitable for studying this research problem.
From among these, limited datasets have been published publicly. \citeauthor{6} (\citeyear{6}) and \citeauthor{9} (\citeyear{9}) provide authorized access to their datasets upon email request. To the best of our knowledge, only \citeauthor{11} (\citeyear{11}) focuses specifically on cartoon videos. 
 However, they only include Elsagate related videos and have not extended their analysis to other types of malicious cartoon content which we have tried to highlight. \citeauthor{4} (\citeyear{4}) address the problem of identifying inappropriate advertisements that are watched by children when shown during kid-appropriate videos; their dataset has a list of inappropriate advertisements instead of videos. We have used their list of appropriate channels to generate seed videos for our dataset. While \citeauthor{1} (\citeyear{1}) and \citeauthor{5} {\citeyear{5}}) share their dataset of malicious and benign videos, they include non-cartoon YouTube videos. For instance news, sports, Hollywood movie clips, etc. are included as part of the dataset as being malicious and not suitable for kids. Furthermore, a number of the videos in these datasets have been removed by YouTube and hence an up-to-date dataset is required. Most importantly, none of the datasets have annotated the existence of absence of specific malicious audiovisual features which have been identified as harmful by psychologists.
\subsubsection{Spatio-temporal Audiovisual Features}
The spatio-temporal, audiovisual features that we have emphasized in our dataset have been left unconsidered whereas their impact on toddlers' and preschoolers' physical and mental development is significant. These features entail both audio and video contents and can be enumerated as follows:
\begin{table}[h!]
\centering
\begin{tabular}{|p{0.45\columnwidth} | p{0.45\columnwidth}|}
 \hline
 \rowcolor{Gray}
 \textbf{Video} & \textbf{Audio} \\ [0.5ex] 
 \hline
 fast repetitive motions & loud music/noise\\ 
 \hline
 scary/disgusting appearance & screaming or shouting\\
 \hline
 hurting/destruction/killing activity & explosion or gunshot sounds\\ 
 \hline
obscene/indecent activity & use of offensive language  \\
 \hline
\end{tabular}
\caption{Malicious  features backed by psychology studies}
\label{tab:features}
\end{table}

\subsubsection{The Collection Process}
Our first step was to create a list of channels for both of our classes: 1) benign and 2) malicious. For benign, we used the channels list from \citeauthor{4} (\citeyear{4}) which comprised of 50 channels. These 50 channels included around 47K videos which were extracted using the YouTube Data API. But as our scope of the problem was cartoon videos only, we performed a random non-exhaustive inspection of around 5 videos from each channel and manually sifted out 16 non-cartoon channels. From the remaining 34 benign channels around 11 videos were selected from each, resulting in 363 videos. For each video we made 10-second clips and randomly selected 5 clips. A few of the videos had less than 5 clips as they were shorter than a minute. We ended up with a total of 594 benign  clips. Similarly, for the malicious class, we started with 32 channels which we gathered from a walkthrough of related videos starting with search results using keywords like ``angry number blocks'', ``angry mario'', ``scary peppa'', etc. Although there were around 14K videos extracted from this seed pool of channels, we selected top 680 videos according to the view count and ended up with 1281 10-second clips of malicious videos. The statistics of seed videos before annotation are summarized in Table \ref{tab:stats1}. Table \ref{tab:stats2} includes the dataset details after the annotation was performed on the seed videos. We excluded anime clips for consideration as well as videos that are neither cartoon videos nor videogame gameplay.

\begin{table}[h!]
\centering
\begin{tabular}{|c|c|c|c|c|} 
 \hline
 \rowcolor{Gray}
 \textbf{Class} & \textbf{Channels} & \textbf{Videos} & \textbf{Clips} \\ [0.5ex] 
 \hline
 Benign & 50 & 363 & 594\\ 
 \hline
 Malicious & 32 & 680 & 1281\\
 \hline
\textbf{Total} & \textbf{82} & \textbf{1043} & \textbf{1875} \\
 \hline
\end{tabular}
\caption{Statistics of MOB dataset seed videos.}
\label{tab:stats1}
\end{table}

\begin{table}[h!]
\centering
\begin{tabular}{|c|c|} 
 \hline
 \rowcolor{Gray}
 \textbf{Class} & \textbf{Clips} \\ [0.5ex] 
 \hline
 Benign & 1046\\ 
 \hline
 Malicious & 519\\
 \hline
 Excluded & 310\\ \hline
\textbf{Total} & \textbf{1875} \\
 \hline
\end{tabular}
\caption{Statistics of annotated MOB dataset.}
\label{tab:stats2}
\end{table}

\subsubsection{Data Preparation}
After collecting the list of video-IDs of YouTube videos, we downloaded the videos using the pytube API at a frame rate of 25. Then for each video we made clips of duration 10 seconds using the moviepy API which is a wrapper for the ffmpeg framework. The nomenclature for a video clip \# x for a video-id y was kept as y\_x. Similarly, we selected 5 random clips from the set of clips generated from a video and for each of the clips extracted frames. As the frame rate was 25, each of the 10-second clip resulted in 250 frames. Each frame was saved in a directory bearing the name of the clip, where the frame names were simply the frame numbers ranging from 0-249 but padded to string of length 4 where \textit{0000.jpg} being the first frame's name whereas \textit{0249.jpg} being the last. These frames were provided as input to the different benchmarks.

\subsubsection{Ethics}
Another principle of law is ``fair use'' which in special cases allows others to use existing artifacts that have ownership protection, without the explicit consent of the owner (\cite{fairuse}). Within the US, research is one of the tasks taken into consideration as fair use (\cite{fair}).  We did not gather any user's web usage activity, personal information or any kind of demographics for YouTube, YouTube Kids, or for the annotation tool. For the annotation tool web application, we generated anonymous login credentials for each user and hence the annotator cannot be identified.

\subsection{Annotation Process}\label{AT}
Our MOB dataset was annotated using a web-application, MAWA (MOB Annotation Web Application), which was custom-built for annotating malicious videos. There is a user authentication feature to keep track of progress of each user's annotation. However, to ensure privacy of users, we created generic credentials e.g. user1, user2, and so on. Each user was randomly assigned a pool of 125 unannotated clips from the database. The user interface is very user-friendly and easy to use with negligible training time. Users can perform annotations over multiple sessions at their convenience. The user interface is inspired by the interface of \citeauthor{kinetics} (2017) which was used to annotate human actions.
\subsubsection{Annotation Methodology}
\begin{figure}[htp]
    \centering
    \includegraphics[width=0.9\columnwidth]{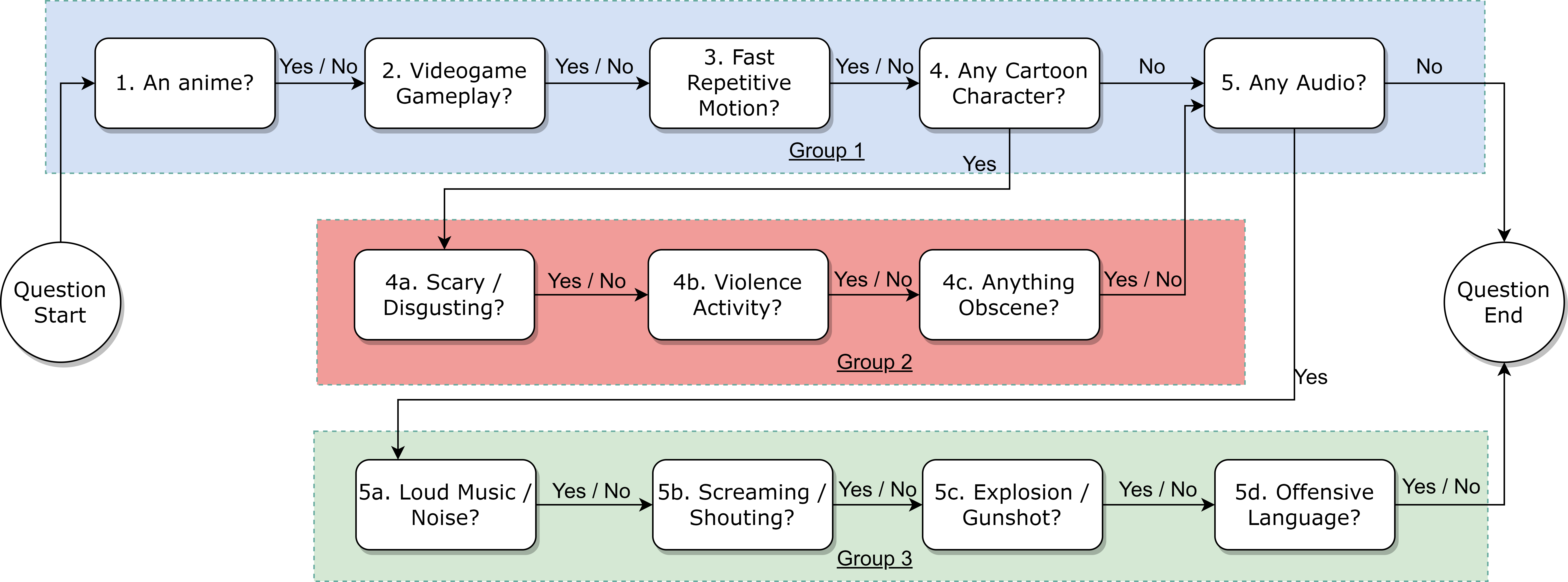}
    \caption{Annotation questions grouped together for fast annotation}
    \label{fig:annotationflow}
\end{figure}

We enumerated the list of all possible malicious video and audio features which are unsuitable for kids of ages 1-5 to watch, as shown in Table \ref{tab:features}.   For every video clip, we formulate questions based on the features and ask the annotator whether they noticed the presence of that feature. From these set of questions we further made groups as shown in Figure \ref{fig:annotationflow} following the effective strategy of multi-class annotation as proposed by \citeauthor{stanford_annotation_paper} (\citeyear{stanford_annotation_paper}). Using this strategy we have split our question set into 3 hierarchy groups:
\begin{itemize}
    \item Group 1: to find the category of the video clip (cartoon, videogame gameplay, anime, or non-cartoon)
    \item Group 2: only asked if the video clip is a cartoon
    \item Group 3: only asked if there is audio in the clip
\end{itemize}
Questions in Group 1 belong to top hierarchy level and will be asked for each video clip. However, if there is no cartoon character in the video clip, Group 2 will be skipped as it is not categorized as a cartoon video. Similarly, if there is no audio in the video clip being annotated none of the questions in Group 3 will be asked from the annotator.
Each video is viewed by one annotator; as each question depicts one feature this makes the questions simple and easy to comprehend, hence not requiring the annotator to make any inferences. Therefore, the possibility of the annotation responses being highly subjective is negligible. However, we did perform a random validation cycle after the annotation phase in which random videos were selected from the set of annotated video clips and one user performed annotation on those again. The validation accuracy was more than 90\%. 

\subsubsection{Annotation Tool}
\begin{figure}[htp]
    \centering
    \includegraphics[width=\columnwidth]{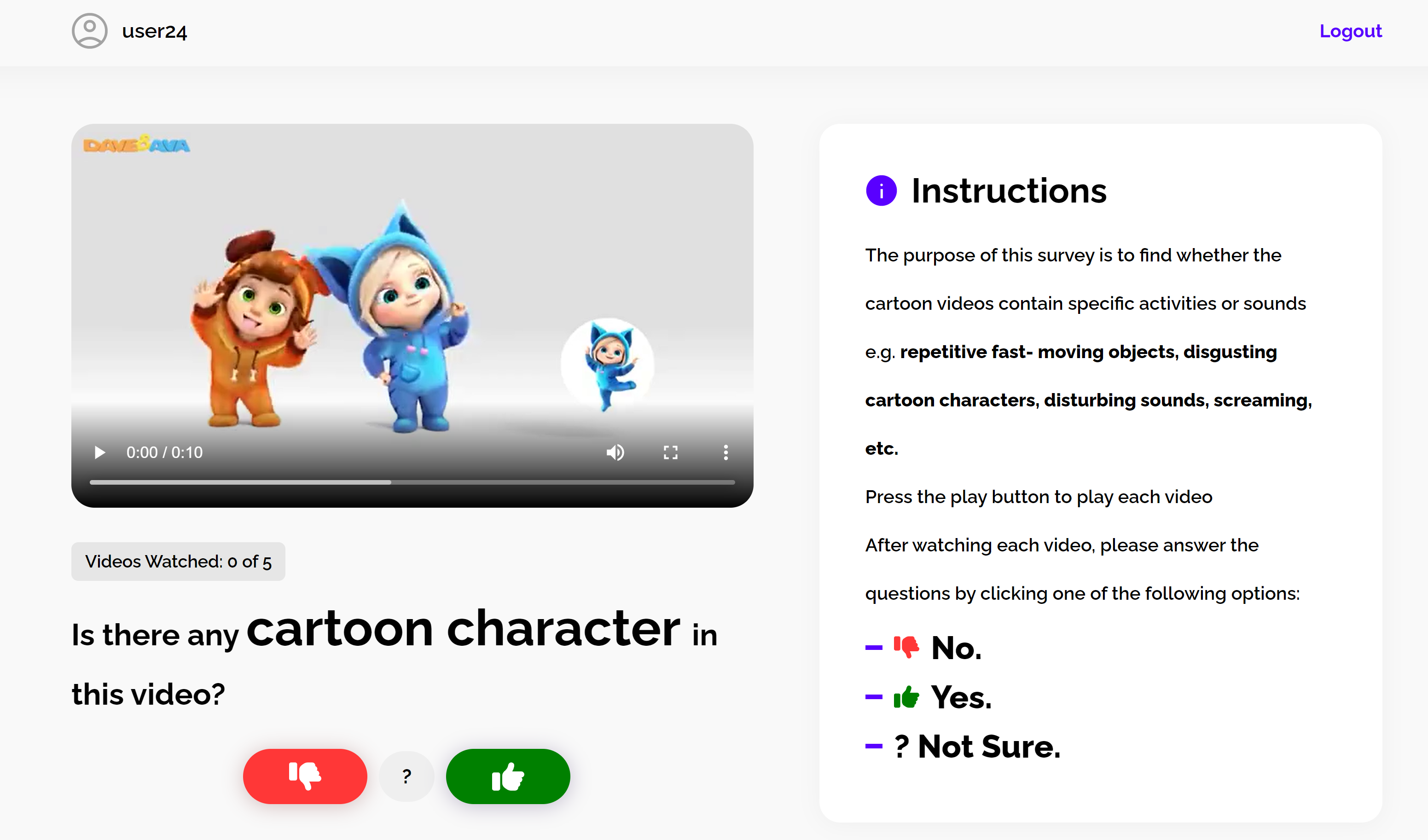}
    \caption{Layout of MAWA (MOB Annotation Web Application)}
    \label{fig:annotationtool}
\end{figure}

The MOB Annotator Web Application (MAWA) was specially developed considering the intricate requirements of the dataset annotation process. We used the following development tools and technologies: HTML, CSS, Javascript Bootstrap library, FASTAPI (Python) for Web services, and MySQL database. Each user is authenticated before the annotation tasks. For each feature's annotation, the response is saved to the database server and hence allows user to restart if the user wishes to complete the task in multiple sessions. The source code of the current version will be made available with easy set-up instructions on how to configure and run the web application. 
\\
MAWA comes with a highly intuitive graphical interface as shown in Figure \ref{fig:annotationtool}. There are flexible controls available for the annotator to play, pause, rewind, view in full screen, and adjust volume for the video clips. Below the video panel the annotator can view the count of total videos annotated as well as the total videos assigned, as well as the questions and buttons to give response. On the right side, is the instruction panel where the annotator can read the comprehensive self-explanatory instructions to become acquainted with the environment and understand what action each button performs.

\begin{figure}[ht!]
    \centering
    \includegraphics[width=0.45\textwidth]{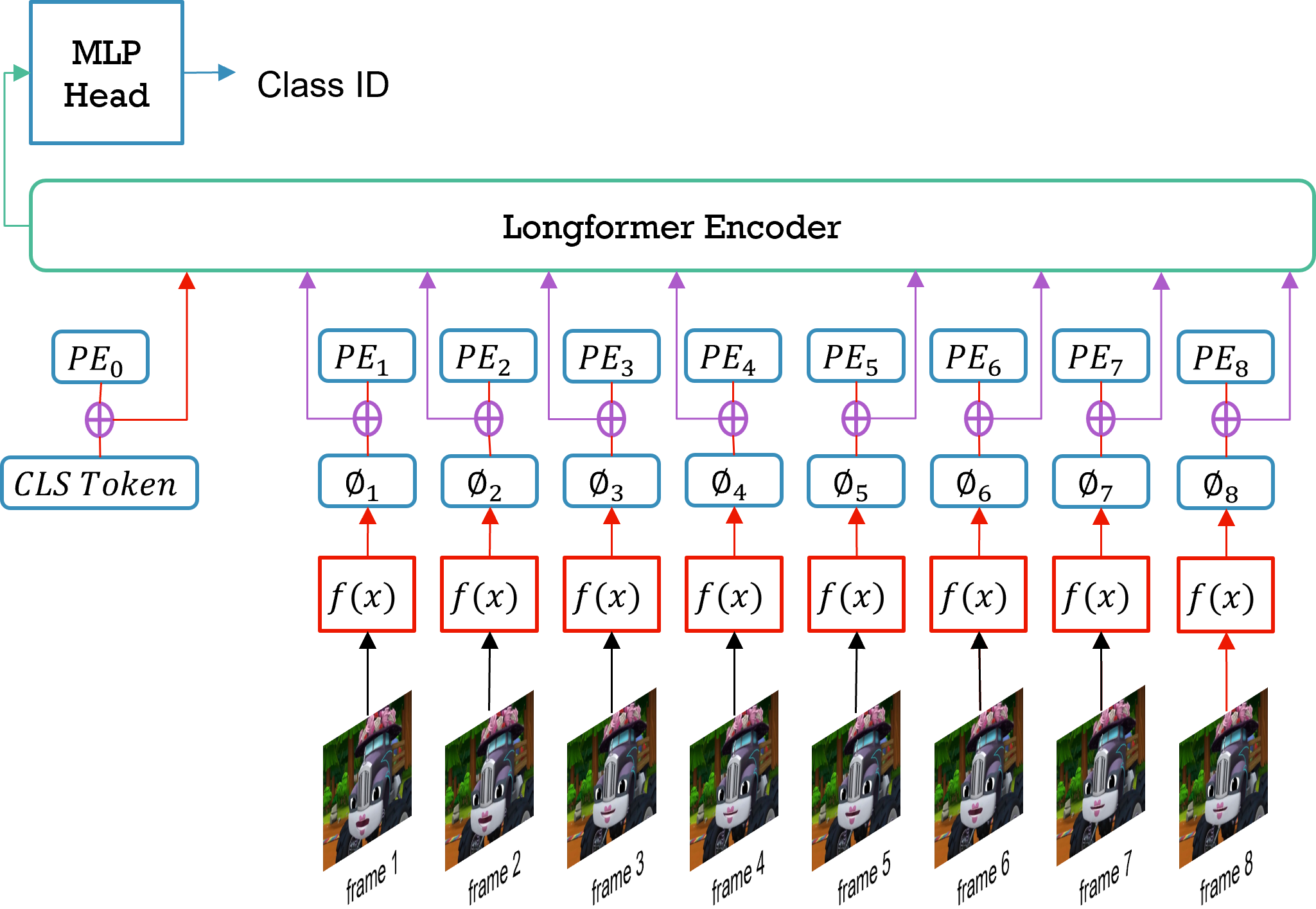}
    \caption{Architecture of Video Transformer Network (VTN) as applied to the frames of our MOB dataset.}
    \label{fig:vtn}
\end{figure}

\begingroup

\renewcommand{\arraystretch}{1.6}
\begin{table*}[!h]
\centering

\begin{tabularx}{\textwidth} { 
  >{\raggedright\arraybackslash}X 
  | >{\centering\arraybackslash}X 
  | >{\centering\arraybackslash}X 
  | >{\centering\arraybackslash}X 
  | >{\centering\arraybackslash}X 
  | >{\centering\arraybackslash}X 
  | >{\centering\arraybackslash}X 
  | >{\centering\arraybackslash}X 
  | >{\centering\arraybackslash}X }
 \multirow{2}{5em}{\textbf{Model}} & \multicolumn{4}{c |}{\textbf{Training}} & \multicolumn{4}{c}{\textbf{Testing}}\\ 
 \cline{2-9} 
 & \textbf{Accuracy} & \textbf{Precision} & \textbf{Recall} & \textbf{F1} & \textbf{Accuracy} & \textbf{Precision} & \textbf{Recall} & \textbf{F1} \\
 \hline
 \hline
 VTN & \textbf{97.93} & \textbf{98.18} & \textbf{97.91} & \textbf{97.86} & \textbf{77.85} & \textbf{82.27} & \textbf{87.17} & \textbf{82.69} \\
 I3D & 93.33 & 95.64 & 96.79 & 96.03 & 72.11 & 80.42 & 82.03 & 80.92 \\
 ConvLSTM & 72.39 & 73.26 & 93.45 & 80.78 & 69.71 & 78.87 & 85.95 & 78.07 \\
 
\end{tabularx}
\caption{Training and evaluation scores for our benchmarks. The transformer-based VTN achieves the best performance.}
\label{table:main_res}
\end{table*}
\endgroup

\section{Benchmark Evaluation}\label{BE}
To stimulate research on video content moderation, we have also included results from a competitive set of video analysis benchmarks.    
This section describes the the deep learning models we selected for our evaluation. We employ three different approaches: 1) first, for the baseline experiment, we consider a 3D convolutional neural network (CNN) namely I3D \cite{i3d}; 2) then we have a recurrent architecture i.e., Convolutional Long Short-term Memory (ConvLSTM) \cite{lstm1,lstm2}; 3) and lastly we adopt a more recent transformer-based approach, the Video Transformer Network (VTN) \cite{vtn}. These approaches are briefly discussed in the following subsections.

\subsection{VTN}

VTN is a deep learning model that utilizes the Longformer \cite{longformer} model which is based on the original Transformer model introduced by \citeauthor{transformer} (\citeyear{transformer}).  It extracts features from a 2D CNN and combines them with positional encoding to learn temporal relationships for video specific tasks such as classification. It is robust to long video sequence, since it is based on Longformer which essentially removes the limitation of the original Transformer and replaces the self-attention mechanism with a more scalable windowed attention along with task specific global attention.

\subsection{ConvLSTM}

ConvLSTMs~\cite{lstm1,lstm2} combines convolution layers with LSTM layers where convolution layers extract features while LSTM layers process that information over multiple time intervals. They have been used in literature for various tasks ranging from video classifications, time series analysis, anomaly detection, and natural language processing tasks. As we are working with video classification task, they were a natural choice for a benchmark.

\subsection{I3D}

I3D~\cite{i3d} has been used in various approaches as both the main network as well as a backbone network for extracting features. In I3D, the pretrained 2D kernels of a deep network are expanded into 3D kernel which helps with learning spatio-temporal features.

\subsection{Training and Evaluation}
We trained each network for 100 epochs with a batch size of 8 for VTN and ConvLSTM while a batch size of 16 was used for I3D. Both VTN and I3D were trained with the SGD optimizer while ConvLSTM was trained with Adam optimizer with a learning rate of $1e^{-3}$, which was reduced every $10^{th}$ epoch, and standard cross entropy loss function. In addition, each model is then evaluated at every $5^{th}$ epoch. All the models were trained on NVidia RTX 3090 GPU. 

Each clip has a length of 10 seconds where each frame has a spatial dimension of $224 \times 224$ and eight frames were used for the temporal dimension. In addition, we also used standard data augmentation techniques such as random cropping, color jitters, horizontal flipping, and normalizing.   

\subsection{Results and Discussion}

\begin{figure*}[htp]
    \centering
    \includegraphics[width=0.5\columnwidth]{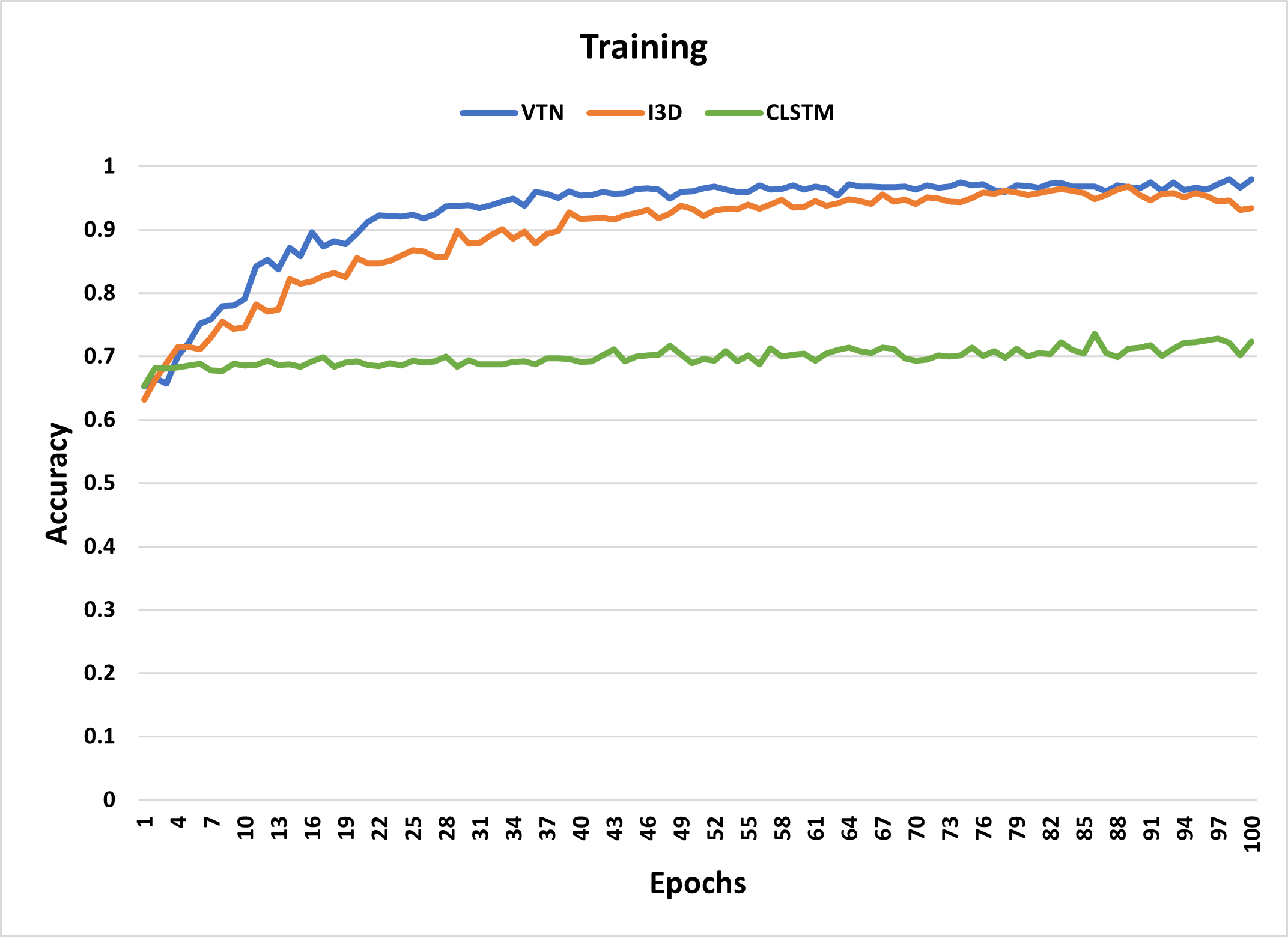}
    \caption{Training accuracy over epochs for the three benchmark models. VTN gives the best performance.}
    \label{fig:training}
\end{figure*}

Our empirical analysis shows that the VTN outperforms both the ConvLSTM as well as the I3D models, with an F1 score of 82.69 on the test set. These results are summarized in Table \ref{table:main_res} and also in Figure \ref{fig:training}. It is unsurprising that the transformer-based approach outperforms the others as they can often learn better spatio-temporal features, can process longer time horizons, and are scalable too, while both ConvLSTM and I3D lack the capability of self-attention and are not as scalable as transformers.  

\subsubsection{VTN Variations}
We evaluated the performance of different variations of the VTN, including modifying the hidden dimension size, the number of transformer encoder layers, and other hyper-parameters. These results are reported in the Appendix.

Note that we have used competitive video classification models as our benchmarks.  None of them are specialized for our task or our dataset.  We hope that they will provide a fair but challenging benchmark for authors interested in doing research on content moderation for children's cartoons.

\section{Conclusion and Future Work}
This paper addresses the problem of effective content moderation for children's cartoon videos. 
 Consumption of malicious content by children between ages 1-5 adversely affects their behavior and cognitive development. We enumerated a set of fine-grained malicious audiovisual features, and also presented the associated medical studies and findings supporting their exclusion. This paper proposed a comprehensive solution including a dataset, a customizable annotation tool for videos, and a benchmark suite of state-of-the-art video classification models. The dataset and annotation tool will be made available for others to stimulate research in this area.
 
In future work, we will focus on analyzing the disturbing audio elements present in our dataset as well leveraging meta-data, subtitles, and user comments within a multimodal classification framework. We also plan to explore few-shot learning and transfer learning techniques as mechanism to overcome data limitations.

\bibliographystyle{flairs} 
\bibliography{references.bib}
\clearpage
\onecolumn
\section*{Appendix}
This section includes additional examples from our dataset and more information about the settings used to train our machine learning models.

\begin{figure*}[!h]
  \begin{subfigure}{.42\textwidth}
    \centering
    \includegraphics[width=\linewidth]{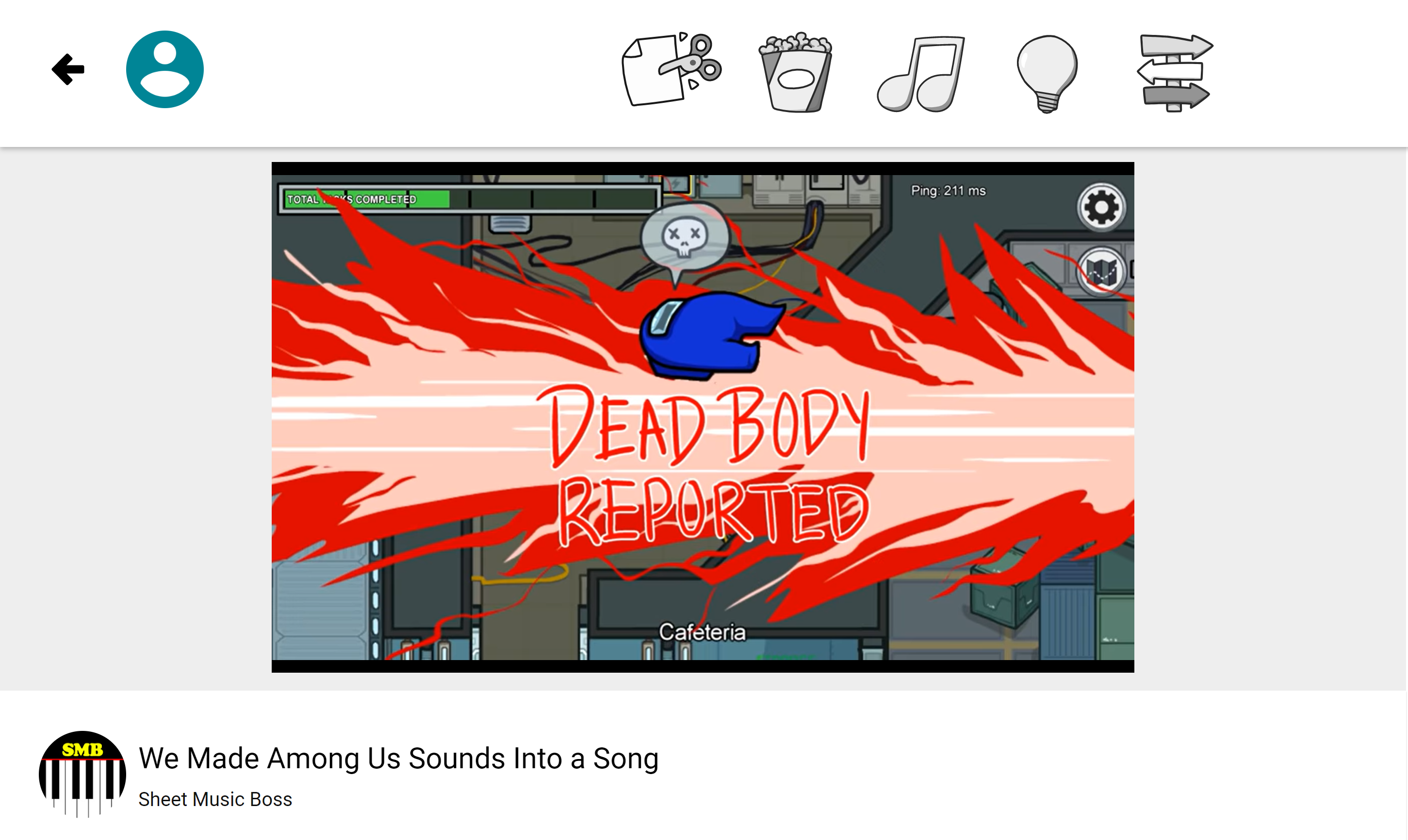}
    \caption{ }
  \end{subfigure}
  \hfill
  \begin{subfigure}{.42\textwidth}
    \centering
    \includegraphics[width=\linewidth]{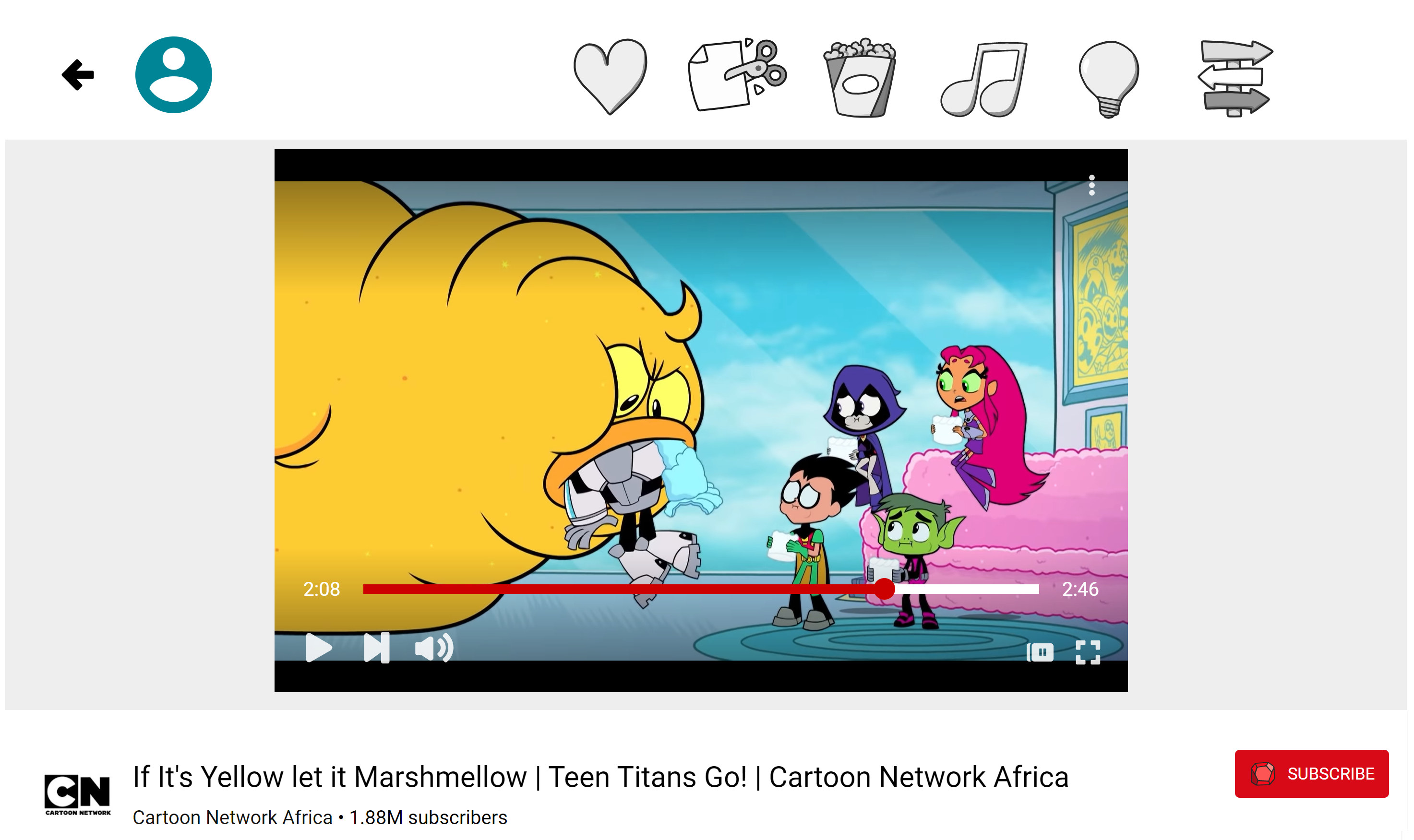}
    \caption{ }
  \end{subfigure}

  \medskip

  \begin{subfigure}{.42\textwidth}
    \centering
    \includegraphics[width=\linewidth]{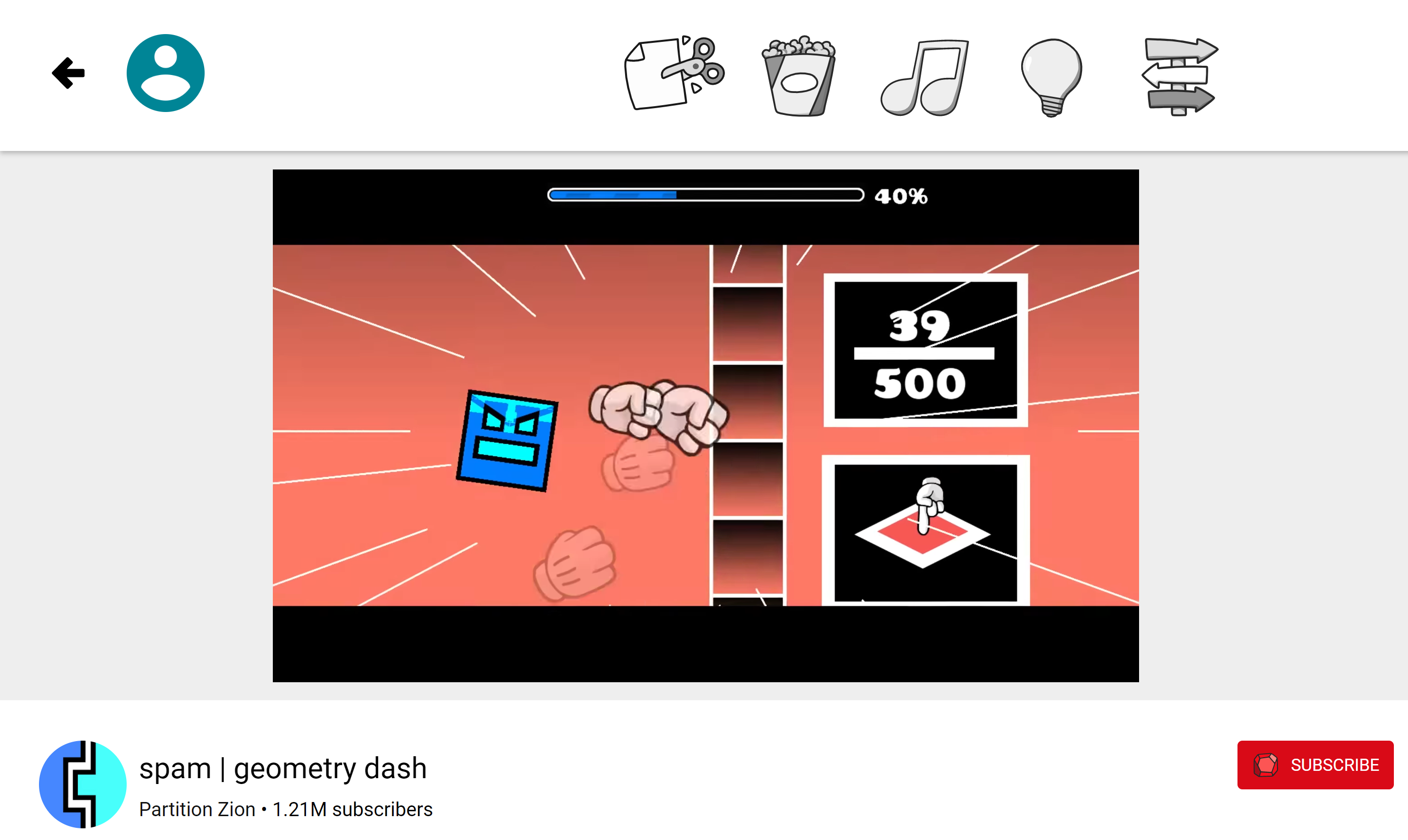}
    \caption{ }
  \end{subfigure}
  \hfill
  \begin{subfigure}{.42\textwidth}
    \centering
    \includegraphics[width=\linewidth]{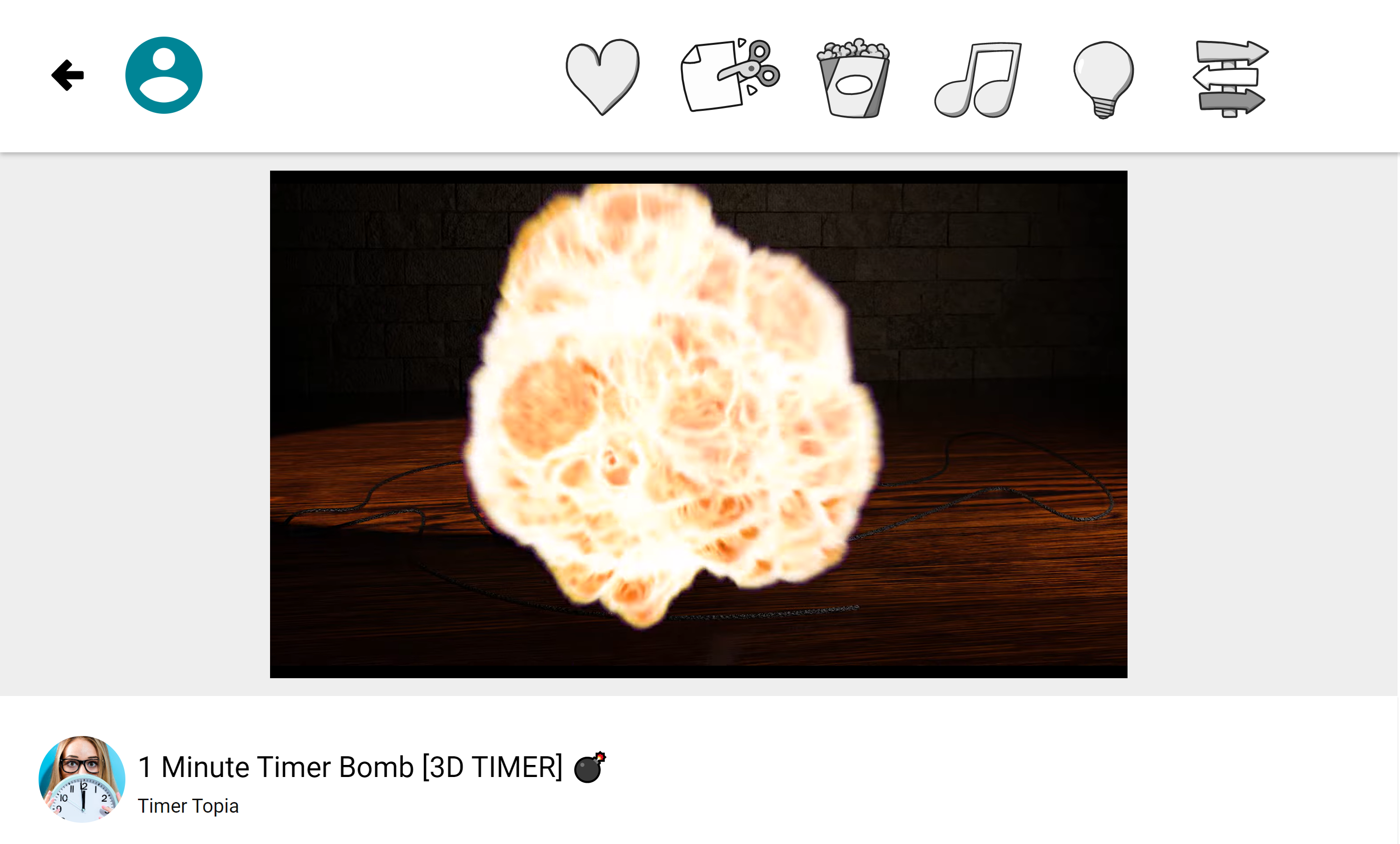}
    \caption{ }
  \end{subfigure}
  \caption{Few snapshots of malicious videos still viewable on YouTube Kids. (a) Among Us game (rated inappropriate for kids under 10 by ESRB) related video where there is loud music and sounds as well as a display of violence in the form of a dead body. (b) Disturbing violent activity where one of the cartoon characters is eating the other cartoon character. (c) Cartoon character aggressively punching the block wall. This video also contains fast moving visuals as well as fast-paced music. (d) This video shows a time bomb which explodes at the end.}
\label{fig:fig1}
\end{figure*}

\begingroup
\renewcommand{\arraystretch}{1.25}
\begin{table*}[!h]
\centering

\begin{tabularx}{0.8\textwidth} { 
  >{\centering\arraybackslash}X 
  | >{\raggedright\arraybackslash}X 
  | >{\raggedright\arraybackslash}X
  | >{\raggedright\arraybackslash}X }
 
 \textbf{Variations} & \textbf{Hyper-parameters} & \textbf{Training Scores} & \textbf{Testing Scores} \\
 \hline
 \hline
 Variation 1 & {Batch Size: 8 \newline \#Heads: 4 \newline \#Layers: 2 \newline Hidden Dim: 2048} & {Precision: 72.97 \newline Recall: 90.96 \newline F1: 79.54 \newline Accuracy: 72.86} & {Precision: 70.98 \newline Recall: 92.07 \newline F1: 81.03 \newline Accuracy: 70.58}   \\
 \hline
  Variation 2 & {Batch Size: 8 \newline \#Heads: 4 \newline \#Layers: 3 \newline Hidden Dim: 2048} & {Precision: 96.87 \newline Recall: 97.06 \newline F1: 96.67 \newline Accuracy: 95.77} & {Precision: 81.23 \newline Recall: 85.98 \newline F1: 82.42 \newline Accuracy: 76.21}   \\
 \hline
  \textbf{Variation 3} & {Batch Size: 8 \newline \#Heads: 6 \newline \#Layers: 3 \newline Hidden Dim: 3072} & {Precision: 98.18 \newline Recall: 97.91 \newline F1: 97.86 \newline Accuracy: 97.93} & {Precision: 82.27 \newline Recall: 87.17 \newline F1: 82.69 \newline Accuracy: 77.85}
 
\end{tabularx}
\caption{VTN Variations. Effects of various hyper-parameters on VTN performance. Variation 3 is the variation used in the paper.}
\label{table:vtn_var}
\end{table*}
\endgroup

\end{document}